\documentclass[letterpaper, 10 pt, conference]{ieeeconf}  %

\IEEEoverridecommandlockouts                              %

\title{\LARGE \bf
No-frills Dynamic Planning using Static Planners
}

\author{Mara Levy, Vasista Ayyagari, and Abhinav Shrivastava\\[0.5em]
University of Maryland, College Park
}

\usepackage{graphicx}
\usepackage{amsmath, wrapfig}
\usepackage{dsfont}
\usepackage{xcolor}

\usepackage{amsmath}

\DeclareMathOperator*{\argmin}{arg\,min}

\makeatletter
\let\NAT@parse\undefined
\makeatother
\usepackage[numbers,sort,compress,square]{natbib}
\usepackage{hyperref}
\hypersetup{
    colorlinks = true,
    citecolor = blue,
    urlcolor = blue,
}
\begin{document}

\maketitle
\thispagestyle{empty}
\pagestyle{empty}

\begin{abstract}

In this paper, we address the task of interacting with dynamic environments where the changes in the environment are independent of the agent. We study this through the context of trapping a moving ball with a UR5 robotic arm. Our key contribution is an approach to utilize a static planner for dynamic tasks using a Dynamic Planning add-on; that is, if we can successfully solve a task with a static target, then our approach can solve the same task when the target is moving. Our approach has three key components: an off-the-shelf static planner, a trajectory forecasting network, and a network to predict robot's estimated time of arrival at any location. We demonstrate the generalization of our approach across environments. More information and videos at {\normalfont \href{https://mlevy2525.github.io/DynamicAddOn/}{https://mlevy2525.github.io/DynamicAddOn/}}.

\end{abstract}

\section{INTRODUCTION}

Humans have an innate ability to understand the world around us. We do not have to think twice when we dodge people on the street or prevent something from falling off the table. Robots, on the other hand, are relatively inept at performing the same tasks. They cannot adapt to dynamic situations in the same way people do. In the past, robots have been utilized in specialized environments, where assumptions can be made about the surrounding environment and the object location. The goal of this project is to leverage recent advancements in robot learning, path planning, and trajectory forecasting to enable robots to act in dynamic environments.
	
In a typical robotic planning problem, the robot observes the environment, plans to complete a specific goal, and then executes one step of that plan, repeating this process until the goal is met. However, in a dynamic environment, accomplishing this is infeasible -- as a robot plans to meet a specific target, the real target will have already moved. Employing a typical planning approach to solve this (e.g., pure pursuit control) results in a robot following behind the target, often without reaching it. In this work, we will address this problem by planning for a possible future location of the object and then adjusting this plan as necessary, to reach the final target position \emph{at the same time as the actual object}.
	
The field of dynamic targets is relatively under-explored, other than tasks like catching a ball~\cite{5980114, 10.5555/591553.591909, Zeng2019VisualRL, 6385963, Kim2014CatchingOI}. However, the task of catching does not require the exact timing; for example, an agent can go to the goal location early and wait for the object. Compare this to our setup of trapping a rolling ball with a box. If the robot reaches the location early, it misses the ball entirely. We demonstrate that synchronizing the agent's timing and the moving object is critical in completing such dynamic tasks. With the notable exception of~\cite{Zeng2019VisualRL}, most other approaches for solving dynamic tasks rely on solving for the physics that represents an object, which fails to generalize beyond the specific task these equations are designed for.

Solving the simple trapping task, raises several research challenges. The robot needs to know a goal location to go to, but predictions of object movements in the real-world are noisy. The standard solution of re-planning needs to work in tandem with the future location predictions. The speed at which we get the predictions needs to be in sync with the speed at which the robot needs them. The approach needs to be real-time, and therefore should use as little computation as possible. The forecasting approach should implicitly learn the underlying physics in the environment, such as the initial velocity of the object and friction, which dictate the movement of objects. Finally, the solution should not require re-training for every possible dynamic setup.

Our key contribution is an approach to utilize a static planner for dynamic tasks using a \textbf{Dynamic Planning add-on}; i.e., if we can successfully solve a task with a static target, then our approach can solve the same task when the target is moving. Our planner integrates a given static planner with two new networks, a trajectory forecasting network, and a robot's estimated arrival network. Using these three networks, our planner is able to determine an appropriate goal location and roll out actions that the robot should take at each step of an episode to interact with a dynamic object. We demonstrate the effectiveness of our approach on the task of trapping a moving ball and dynamic FetchReach~\cite{fang2018dher}.

\vspace{-0.02in}
\section{RELATED WORKS}
\label{sec:related work}
\vspace{-0.02in}

\noindent\textbf{\textit{Motion forecasting:}} Forecasting the trajectory of an object to improve robotic manipulation has been studied for decades (e.g., at least as early as 1995~\cite{370380}). Since then, the interplay between prediction and robotics has been studied in many different scenarios. In recent years, a popular paradigm to study trajectory forecasting is social situations with many moving actors~\cite{7780479, 10.1007/978-3-030-11015-4_16}. Most of these approaches build on Recurrent Neural Networks, or variants, to predict actors' future movements. Generally speaking, the input and output sequences are of the same short length, limiting their applications in robotics problems, which require long-horizon predictions. Such approaches require a prohibitive amount of memory and time to make accurate long-horizon predictions.
    
A popular use of trajectory forecasting in robotics is predicting how a robot will interact with objects to perform specific tasks~\cite{DBLP:journals/corr/FinnGL16, DBLP:journals/corr/FinnL16, Fragkiadaki2016LearningVP}. However, much of this comes down to learning, or modeling, the physics of an object and the robot, and predicting how the objects will interact with each other. To accomplish these tasks, the algorithms must simulate thousands of possible futures to find the right one. This makes them unlikely to work in environments with dynamic objects, because by the time an action is decided on, the required interaction will likely need to be different.
    
Due to the issues highlighted above, many recent robotics works have resorted to using physics to accomplish their tasks~\cite{5980114, 10.5555/591553.591909}. For the most part, current techniques require large amounts of domain-specific data pre-processing and post-processing, to manually extract velocity, acceleration, position, etc.~\cite{6583227}. Such processing may require information about the environment, that ideally, we would want to learn.

\noindent\textbf{\textit{Dynamic tasks:}} There are a few recent works that address the task of moving targets. In ~\cite{fang2018dher} authors take the original OpenAI Gym environments and develop an end-to-end solution, where a reinforcement learning model tries to hallucinate the future. However, they move the objects along a straight predictable path, not taking into account any physics. Additionally, \cite{KOC2018121, 10.1145/3334480.3382853} address playing ping pong and catching a ball; but as discussed above, they both rely on accurate physics modeling to plan future paths. 
    
Similar to our approach,~\cite{Dosovitskiy2017LearningTA,Zeng2019VisualRL} solve for dynamic goals using future prediction. However,~\cite{Dosovitskiy2017LearningTA} predicts the future, taking both actions and object movement into account, as opposed to just the object movement. This exponentially increases the number of possible futures and adds complexity, without a clear benefit.~\cite{Zeng2019VisualRL} only addresses the problem of catching, which doesn't require as accurate timing as our task. Additionally, they feed their entire future prediction into the action policy. This implies that not only do they have to run their forecaster at each step, which is computationally expensive, but for each task, they have to train both the forecaster and the model-predictive controller; making it difficult to evaluate their generalization to other tasks. We decouple these two tasks, which leads to a more general solution that can be applied to many different problems.

\section{APPROACH}
Our key contribution is a mechanism to retool a static planner using a Dynamic Planning add-on. This implies that if you can successfully solve a task with a static target, then you can solve the same task if the target is moving. A task-centric interpretation would be that our proposed add-on `reduces' a dynamic task to static task, which can be solved by off-the-shelf static planners. 

Our proposed add-on, Dynamic Planner, has three main components that work in tandem to complete a dynamic task. First we describe the dynamic task used in this paper (Section~\ref{sec:task}), followed by the building blocks of our model (Sections~\ref{sec:static},~\ref{sec:trajectory},~\ref{sec:eta}), and how they are integrated to work together (Section~\ref{sec:dynamic}).

\begin{figure*}
    \centering
    \includegraphics[width=0.8\linewidth]{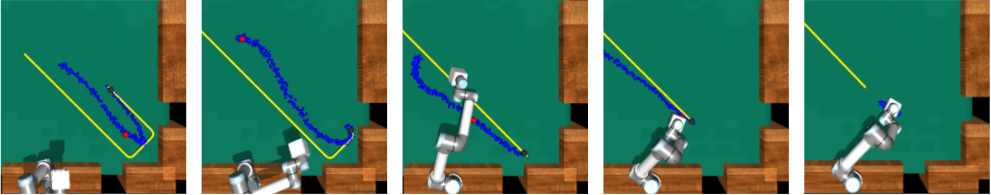} 
    \vspace{-0.1in}
    \caption{Our goal in the Trapper task is to train the robot arm to trap the ball, shown in grey, while it is still moving. The velocity and direction of the ball are randomly set at the beginning of each episode, as is the starting location of the ball. The yellow line represents the known trajectory of the ball, which can be found by running the same episode twice, and the blue line represents the predicted trajectory given only past locations. The arm utilizes the predicted trajectory when selecting a future location, shown in red, to target.}
    \vspace{-0.25in}
    \label{fig:Rollout Example}
\end{figure*}
\subsection{Task definition}
\label{sec:task}
The goal in our task, \textbf{Trapper}, is to trap a moving ball using a robotic arm, visualized in Figure~\ref{fig:Rollout Example}. To be successful, the agent needs to correctly synchronize the time when the attached box covers the goal location and when the ball is at that specific location. If the box traps the ball, by covering it, the episode is considered a success. We chose the UR5 robot because of its widespread use and several successful sim2real demonstrations~\cite{DBLP:journals/corr/abs-1903-07740, Strudel2019LearningTC}. We use a box attached to the UR5 arm, instead of a gripper, to focus on the dynamic aspect of the Trapper task, and omit the added difficulty of getting the gripper in \emph{just} the right position for manipulation. We used the MuJoCo~\cite{conf/iros/TodorovET12} physics simulator to model Trapper.

At the start of each episode, the ball is placed at a random location on a billiard table. A total of eight pockets imply that the ball can roll off the table at any point. This forces the robot to trap the ball while it is still moving, as opposed to the shortcut of waiting for the ball to come to a halt before attempting to trap it. In addition to considering the ball rolling off the table, the planner must learn to predict and deal with the ball bouncing off the walls of the table. Across episodes, we randomize the starting location and initial velocity of the ball. Also, during some test episodes, we change the friction of the table.

Next, we describe how we utilize the static planner for the Trapper-Static task (trap a ball that is not moving) to solve the Trapper-Dynamic task (trap a rolling ball).
\vspace{-0.075in}
\subsection{Overview}
Our approach builds on three building blocks: a static planner, a trajectory forecaster, and a network that predicts how long it takes the arm to reach any particular goal from its' current location (or an estimated time of arrival network). We first describe each of these components and then present how they interact with each other to solve dynamic tasks.

\subsubsection{Static Planner}
\label{sec:static}
Our approach utilizes a static planner. To demonstrate the different possible applications for our algorithm, we use two types of static planners, utilizing model-free reinforcement learning and inverse kinematics, to plan a trajectory towards the goal location.

\noindent\textbf{\textit{Model-free Reinforcement Learning Planner (RL):}} We use Deep Deterministic Policy Gradients~\cite{journals/corr/LillicrapHPHETS15} with Hindsight Experience Replay (HER)~\cite{DBLP:journals/corr/AndrychowiczWRS17}, to solve for the static planner for the Trapper-Static task. Our policy maps the state space $S$ and the goal space $G$ of the environment to an action: $\pi : S, G \rightarrow A$. We will designate the states corresponding to the robot arm as $s^R$ and the current state of the object as $s^O$. Given a static goal, $g\in G$, our policy for the arm is $\pi_{g}(s^R)$, for any $s^R \in S$. In Section~\ref{sec:dynamic}, we will describe how this policy can be utilized to solve dynamic tasks.

We define the reward space used to obtain $\pi_g(s^R)$ as $Q^{\pi_g}(S,A)$. HER typically works best with a sparse reward, $Q^{\pi_g}(S, A) = 1$ if we have reached goal state, $g$, and $Q^{\pi_g}(S, A) = 0$ otherwise. However, in the specific case of the Trapper task, it is difficult for the arm to learn how to align the box horizontally with the table to trap the ball without an additional reward. We address this by solving for a polynomial reward $R(\theta_{\text{box}})$, where $\theta_{\text{box}}$ is the angle between the surface normal of the open-face of the box and the surface normal table. We boost this reward by multiplying by a function $\beta$, which increases as the box nears the ball. This ensures that the box is in the right orientation to trap the ball. In addition, we give a sparse reward based on whether or not the ball is successfully trapped. Therefore, our final reward function is:
\begin{equation} 
Q^{\pi_g}(s^R,a) = \left[\mathds{1}\left(s^R(\text{box}) == g\right) +  \beta R(\theta_{\text{box}})\right],
\end{equation}
where $s^R(\text{box})$ is the location of the box and $==$ checks if the box has reached the goal, with some margin of error.

\begin{figure*}
    \centering
    \vspace{0.05in}
    \includegraphics[width=.95\linewidth ]{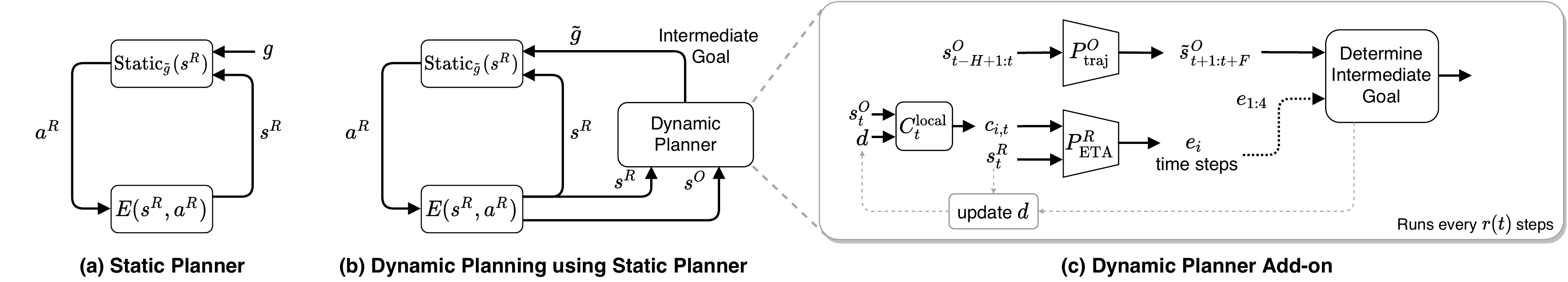} 
    \vspace{-0.1in}
    \caption{\textbf{Planner Overview: }(a) How a simple planner would be run in a static situation. At each step the planner, $\text{Static}_g$ $(\pi_g$ or $IK_g$), takes in $s^R$, the state of the robot, as well as $g$, the goal location. It then outputs $a^R$, the action for the robot to take in the environment. The action is simulated in $E$ and the new state is outputted. (b) The Dynamic Planner is integrated with the static planner, taking in $s^R$ as well as $s^O$, the state of the object and outputting the $\tilde{g}$ to be used in the static planner. (c) The Dynamic Planner is extended to show how $P_{\text{ETA}}^R$, the ETA network for the robot, and $P^O_{\text{traj}}$, the trajectory forecasting network are used to find $\tilde{g}$.}
    \label{fig:Model Overview}
    \vspace{-0.22in}
\end{figure*}

We use the Stable Baselines~\cite{stable-baselines} implementation of DDPG~\cite{journals/corr/LillicrapHPHETS15} and HER~\cite{DBLP:journals/corr/AndrychowiczWRS17} to train this static planner on the Trapper-Static task. After training, this planner achieves $\sim75\%$ success rate on the Trapper-Static task. Failure cases typically occur if the ball is too close to the arm's base or out of the arms reach. We do not re-train this static planner when adapting it for dynamic tasks. Figure \ref{fig:Model Overview}(a) illustrates a static planner during inference. Since we use an off-the-shelf static planner we refer the reader to~\cite{DBLP:journals/corr/AndrychowiczWRS17, stable-baselines} for more details.

\noindent\textbf{\textit{Inverse Kinematics Planner (IK):}} For the second static planner, we use the UR5 inverse kinematics and the dampened least squares method to solve for the trajectory of the arm~\cite{Control}. Ideally, when using an inverse kinematics planner you would want to plan the entire arm trajectory to the goal each time the goal location changes, however this is too time consuming for a dynamic problem. To mitigate this we solve for 5 steps of the path at a time. This gives the flexibility to change the goal location as the episode rolls out. We will refer to this planner as $IK_g(s^R)$ where, the planner takes in the state of the robot, $s^R$, and outputs $a^R$, the next action for the robot to take. Unlike the RL planner, this planner reaches $\sim100\%$ success rate.

\subsubsection{Trajectory Forecasting}
\label{sec:trajectory}
Trajectory forecasting has been widely studied in computer vision and autonomous vehicle communities (Section~\ref{sec:related work}). However, many of its applications (e.g.,~\cite{7780479}) do not require the kind of long-horizon real-time forecasting that is needed for dynamic robotics. Most trajectory forecasting techniques use Recurrent Neural Networks (RNN) or variants, which have memory and time limitations for long-horizon predictions. 

Instead, we opt for a faster, albeit potentially less accurate, approach~\cite{10.1007/978-3-030-11015-4_16} that requires little knowledge about the actual environment and tries to predict $F$ steps into the future given $H$ steps of past information, where $F >> H$. Therefore, we sacrifice some accuracy to gain a real-time long-horizon prediction. This is necessary because, unlike in autonomous driving, the episode happens in a very short period of time. We cannot wait to gather a lot of prior experience. In our specific example, we use $24$ steps of prior information to predict $300$ steps of future information. The lower prediction accuracy has modest impact on the success rate at high velocities, but it satisfies the necessary time constraints, making it an attractive choice for this scenario.

Our trajectory forecasting network uses~\cite{10.1007/978-3-030-11015-4_16}, which uses Convolutional Neural Networks (ConvNets). Unlike RNNs, which make sequential predictions, ConvNets can predict long-horizons in a single-shot leading to faster prediction in our setup. Our selection was also, in part, inspired by the findings of~\cite{DBLP:journals/corr/abs-1803-01271}. The state of the object (i.e., its location) at time $t$, is denoted by $s^O_t = (x_t, y_t, z_t)$. At the current time $t$, the network takes $H$ past object locations, ${s}^{O}_{t-H+1:t}$, and the current location as inputs and outputs predictions for future locations for $F$ time steps. This network can be defined as
\begin{equation}
P^O_{traj}({s}^{O}_{t-H+1:t}) = \tilde{s}^{ O}_{t+1:t+F}
\end{equation}
where $\tilde{s}^{O}_{t+1:t+F}$ represents the predicted object locations at $[t + 1, t + F]$ time steps .

\begin{figure*}
    \vspace{0.1in}
    \centering
    \includegraphics[width=.94\linewidth]{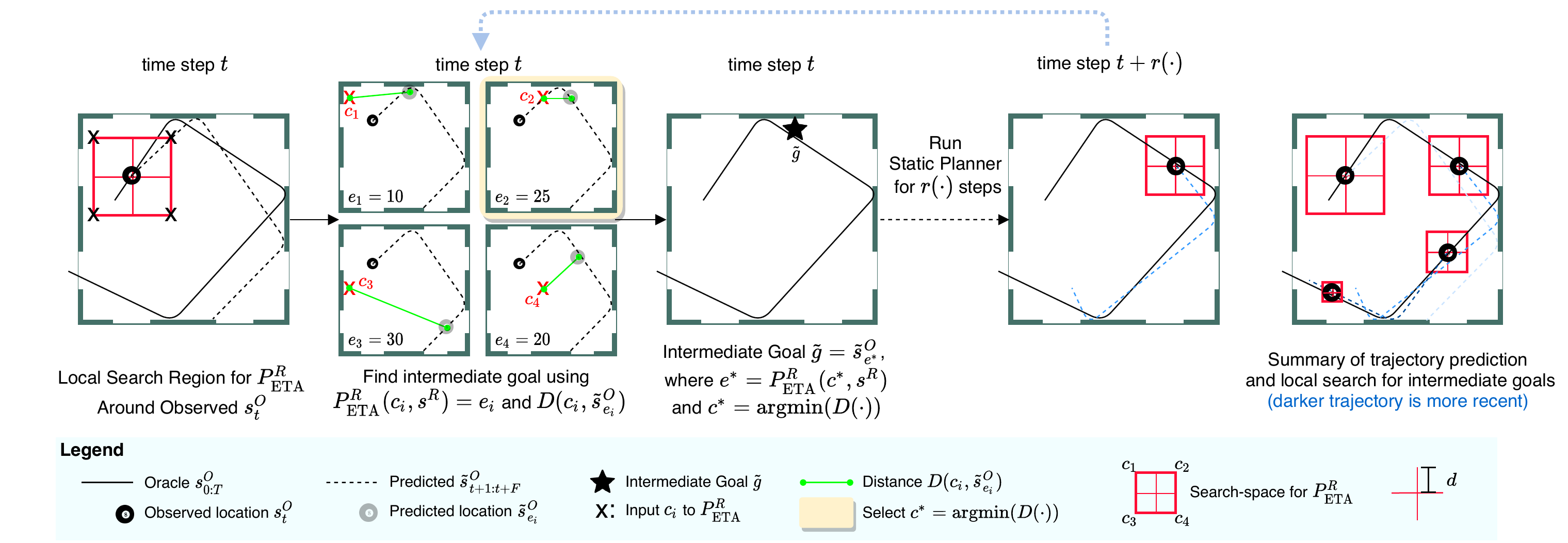} 
    \vspace{-0.1in}
    \caption{The progression of how our dynamic planner selects the new target location.The first image shows the box with edges of size $2d$ drawn. From here we run the ETA network on the corners and find the corresponding points on the predicted trajectory. We select the point on the trajectory closest to the corresponding corner as the new $\tilde{g}$. This is repeated every $t_\text{TTR}$ steps.}
    \vspace{-0.25in}
    \label{fig:Intermediate Goals}
\end{figure*}

These predicted locations are relatively accurate for earlier time steps, but get more inaccurate as the time gets further from $t$. This inaccuracy can be seen in the illustrated dashed blue line in Figure \ref{fig:Intermediate Goals}. However, given the design of our dynamic planner, the predicted trajectory need not be exact as long as it is generally in the right direction. If this condition holds, our algorithm moves the arm in the right direction, so it is always getting closer to the goal location. We quantify this by showing that results using our predicted path are close to the results when we use the oracle future path.

\subsubsection{Estimated Time of Arrival (ETA) of the Arm}
\label{sec:eta}
In previously studied dynamic environments, it did not matter when the robot arrived at the goal location, so long as it did so before the object. For example, when catching a ball as long as the robot is there when the ball reaches the desired location, the robot can successfully catch it. This is also true in other popular dynamic problems, such as ping pong. 

Trapper-Dynamic is different because it requires coordinated arrival timing between the object and the robot. If the robot reaches the target location too early, the box will already be on the table by the time the ball arrives, causing the ball to collide with the box instead of properly trapping it. Likewise, if the robot reaches the target location too late, it will miss its chance to trap the ball entirely. Therefore, the crux of our approach is that we have to time the drop of the box with the exact time the ball will be at the drop location. In order to coordinate this, we define an ETA function $P_{ETA}^R(s^R, g)$, which for every $s^R \in S$ and a given goal location, $g$, provides an estimation for how long it will take our static solver to reach $g$ from state $s^R$.

We use a simple Multi-layer Perceptron (MLP) network to model $P_{ETA}^R$. To get the data to train this network, we run Trapper-Static using the static solver multiple times, for many different $g$, sampling five intermediate states of the robot between the initial state and the goal location. We then calculate how long it takes to reach the goal location from each of those states. We discretize the ETA window of 500 time steps into 100 bins (e.g., 0-4 steps, 5-9 steps, etc.). In order to figure out which of these bins a state falls into we structure our network with 3 linear layers each with a ReLu activation network and a final layer which is also linear and outputs a probability to 100 different buckets using softmax. At training time, we use both the state and the goal as input and classify each pair into one of the bins. We use softmax cross-entropy loss for optimization. We observed that the discretized ETA network led to better convergence than a regression model, and the possible error of up to $4$ time steps for each classification did not have any practical impact.

\subsubsection{Dynamic Planner}
\label{sec:dynamic}

At each time step, our static planner takes in the observed state and the goal, and outputs an action. To utilize this planner for dynamic tasks, we can modify the goal location $g$ input to planner, and there-in lies the challenge. The trajectory forecasting network can give potential future locations of the ball, but its unclear which location should be used as the goal. A simple solution is to use the current location of the ball at every time step $s^O_t$ (i.e., change the goal every time step) and expect that the arm and ball will eventually converge (e.g., pure pursuit control). However, as we demonstrate in experiments, this results in the arm following the ball and is unable to deal with bounces and drop-offs effectively. Another extreme option is to use a long-horizon future ($s^O_T$, where $T >> t$) as a goal and have the arm wait for the ball. However, this assumes precise trajectory prediction, which is erroneous. In our approach, we aim to achieve a balance between these extremes, by utilizing the trajectory forecasting and ETA network together.

Our dynamic planner uses the trajectory network and the ETA network to determine an appropriate goal location, which should ideally reflect the location where the ball and the arm will reach at the same time. However, since the outputs of trajectory network and ETA network are noisy, we need to repeat this process, or re-plan, as time progresses. That said, deciding when to re-plan, is a problem within itself. Ideally, we want to re-plan as little as possible, because it is time-consuming. On the other hand, re-planing means that our initial target location does not have to be accurate, as long as it is in the vicinity of where the ball will be. 

\noindent\textbf{\textit{Re-planning:}} To address this, we propose a simple approach that finds a balance between these two by shrinking the re-plan window each time, that is, as the arm gets closer to the ball, we re-plan more often, allowing the algorithm to zero-in on an accurate target. Our time to re-plan is given by $t_\text{TTR} = \text{max}(\gamma t_\text{TTR}, 25)$. We initially set $t_\text{TTR}$ to be $75$ and $\gamma$ to $0.9$. We don't allow $t_\text{TTR}$ to be less than $25$. This method allows the target to be fairly inaccurate at the beginning of each episode and zero in on the correct location later on. Therefore, our approach determines an appropriate goal location, which we will refer to as intermediate goal $\tilde{g}$, and $t_\text{TTR}$, it then runs the static planner for $t_\text{TTR}$ time steps before it re-plans. This process is illustrated in Figure~\ref{fig:Model Overview}(b).

\noindent\textit\textbf{{Finding intermediate goals:}} The discussion so far alluded that, given the trajectory network and the ETA network, determining an appropriate goal location is straightforward, which is untrue. Next, we describe why this is the case and our approach to determine $\tilde{g}$.

At each re-planning step, the trajectory forecasting network predicts $\tilde{s}^O_{t+1:t+F}$ and we want to find the new value of $\tilde{g}$ from this trajectory such that the following is true, for a current state $s^R$,
\begin{equation} \label{eq:constraint}
\tilde{s}^{O}_{e^*} = \tilde{g} \text{ and } P_{ETA}^R(s^R, \tilde{g}) = e^*,
\end{equation}
that is, both the arm and the ball will reach $\tilde{g}$ in $e^*$ time steps. The problem with these constraints is that they have cyclic dependence, where we need $e^*$, which is some unknown future time step on the trajectory prediction to find $\tilde{g}$, and we need $\tilde{g}$ to find what $e^*$ should be from the ETA network. To resolve this, we could use a brute force approach and get $e_t$ at every time $t$ on the predicted trajectory until we find an $e_t$ that satisfies eq.~\ref{eq:constraint}. However, this is not practical for a real-time setup, at least not without employing a computational cluster. Instead, we devise the following algorithm to approximate $\tilde{g}$.

As illustrated in Figure~\ref{fig:Intermediate Goals}, we begin by considering a local search region (shown as a red box) around the current observed location of the ball $s^O_t$. To determine the size of this box, parameterized by $d$, we solve
\begin{equation}
d = D(\tilde{s}^{O}_{e_t}, s^{O}_{t}) \text{, where } e_t = P_{ETA}^R(s^R, s^{O}_{t})
\end{equation}
where $D$ is the distance function. The intuition is to shrink the local search region proportionally as the arm nears the ball's current location. If the arm is getting closer to the ball, then this number should shrink as the episode progresses. 

Once we have $d$, we take the current location $s^{O}_{t} = (x,y,z)$ and find the four corners of our square search region, located at $(x + d, y + d, z), (x + d, y - d, z), (x - d, y - d, z), (x - d, y + d, z)$. Note that for our current experiment, there is no need to modify $z$ because everything is in on a 2D plane, but it's easy to do so. Let these four locations be $c_1, c_2, c_3, c_4 \in C$. We now solve for $e_i = P_{ETA}^R(s^R, c_i)$ and look at the trajectory to select the following points $\tilde{s}^{O}_{e_i}$. Out of these, we select the point in $\tilde{s}^{O}_{e_i}$ that satisfies \begin{equation}
c^* = \argmin_{c_i} \{D(c_i, \tilde{s}^{O}_{e_i}); \forall c_i \in C\}
\end{equation}
$c^*$ is our best guess at a location close to the actual trajectory where the arm and the ball will reach the same location at the same time. We use $c^*$ to find a good approximation for $e^*$, so we can select a point on our estimated trajectory where we predict the arm and ball will meet at the same time.
This gives us $e^* \approx P_{ETA}^R(s^R, c^*)$. We can determine the intermediate goal as $\tilde{g} = \tilde{s}^O_{e^*}$ from the original constraint, eq.~\ref{eq:constraint}, and pass that into our static planner. Simplified versions of the Dynamic Planner and the intermediate goal finding algorithm are illustrated in Figure~\ref{fig:Model Overview}(c) and~\ref{fig:Intermediate Goals} respectively.

\section{Experiments}
\label{sec:result}

We first present experiments on the Trapper-Dynamic task, described in Section~\ref{sec:task}, and then report results on a dynamic version of the FetchReach environment from~\cite{fang2018dher}.

\noindent\textbf{\textit{Implementation Details:}} We train our model in two separate stages. First, we obtain Trapper-Static using either online reinforcement learning or by solving for the inverse kinematics of the arm. After we have a functioning static model, we simultaneously collect data from MuJoCo for our ETA network and our trajectory prediction network. We do this by using Trapper-Dynamic, but keeping the value of $g$ static, so we can observe both the trajectory of the ball and the path of the arm towards one location. We collected ETA data, as discussed in Section~\ref{sec:eta}. 
In order to get diverse data for the trajectory prediction network, we run each episode four times longer than usual and then sample five random trajectories of length $F+H$.
We use this data to train our ETA and trajectory prediction network offline.
\vspace{-0.075in}
\subsection{Baselines}

\subsubsection{Target Pursuit} This baseline, inspired by pure pursuit control, is the simplest way to solve the dynamic problem. At each frame $t$, we update $\tilde{g}$ to be $s^O_t$.

\subsubsection{Oracle Trajectory} We use an oracle for the trajectory, instead of predicting the trajectory. We obtain the oracle by running the same episode twice. On the first run, we don't move the arm so that we can collect the ball's trajectory with no arm interference. On the second run, instead of feeding in the output of $P^O_{\text{traj}}$ to determine the new $\tilde{g}$, we instead feed in the oracle trajectory. Note that if the arm interacts with the ball (e.g., touches it), then the ground-truth trajectory will be different from the oracle trajectory.

\vspace{-0.075in}
\subsection{Experiment Setup}
We run two different experiments. First, we change the ball's initial velocity across episodes, randomly selecting the x and y components. Second, we change the friction loss of the ball, which is a simple way to increase the friction between the ball and the table. Note that all networks were trained with the first variant, but the second variant is never observed during training, and thus, evaluates generalization.

\subsubsection{Number of episodes and time steps} To obtain a data point, we run our algorithm on $100$ episodes of Trapper-Dynamic with the same initial settings. For each initial setting, we collect 10 data points and plot the average and standard deviation on the resulting graph. We ran all our experiments over $500$ time steps in MuJoCo.

\subsubsection{Metrics and `Solvable' episodes} For each data point, we collect two different metrics: (1) the success rate over all $100$ episodes, and (2) success rate over all `solvable' episodes of those $100$ episodes. To determine if an episode is solvable, we use an oracle ETA network, which knows the exact amount of time it takes from the initial arm location to reach every point on the table. We first get the oracle trajectory, and then, observe if the arm can reach any point on the trajectory before the ball reaches the same point. If this is possible we say the episode can be solved, that is an episode is solvable if $e_t\le t$, where $e_t = P_\text{ETA}^R(s^R, s^O_t)$.

\begin{figure}[t]
\vspace{0.05in}
    \centering
    \includegraphics[width=.48\linewidth]{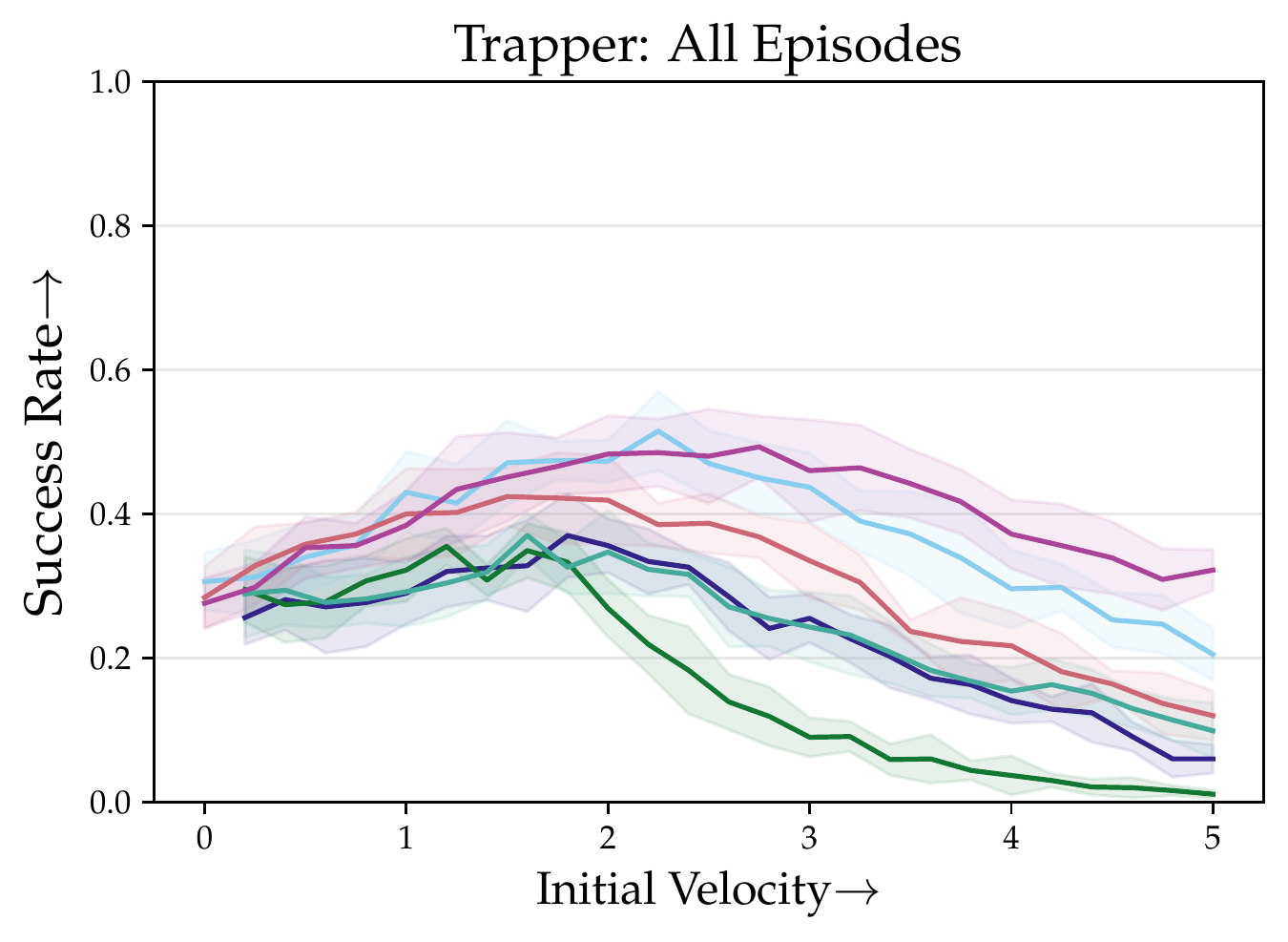} 
\includegraphics[width=.48\linewidth]{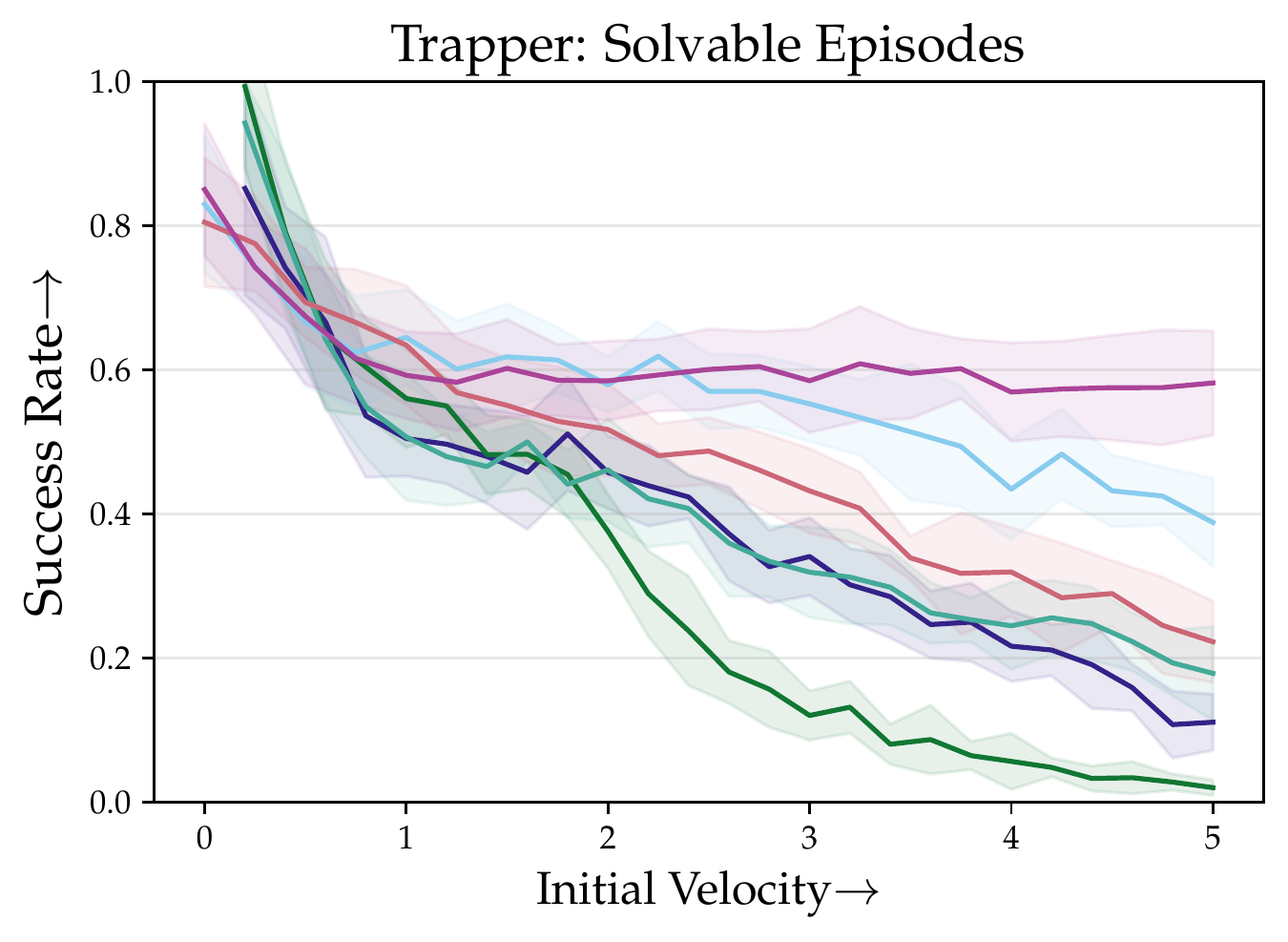}
\includegraphics[width=.48\linewidth]{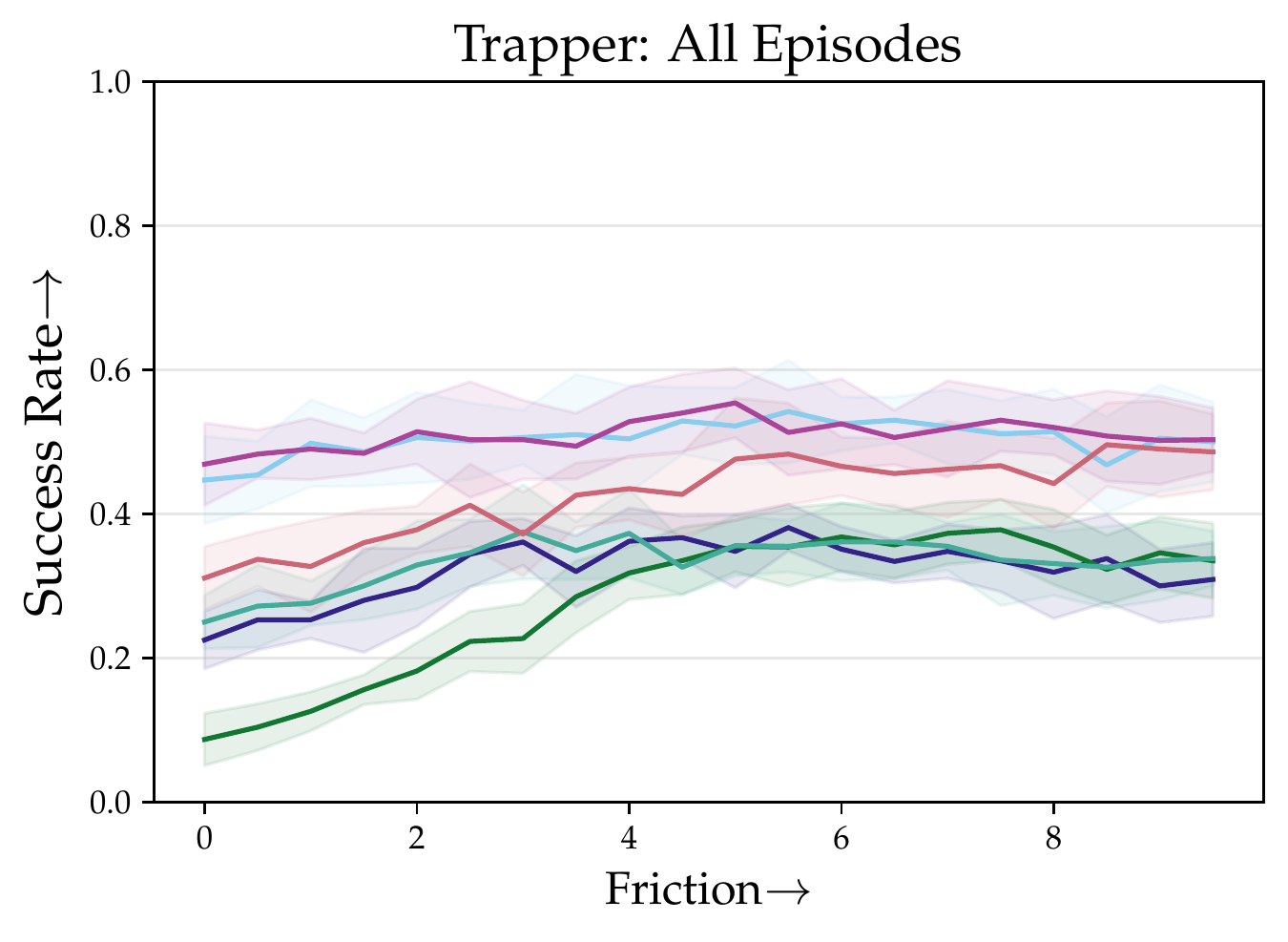}
\includegraphics[width=.48\linewidth]{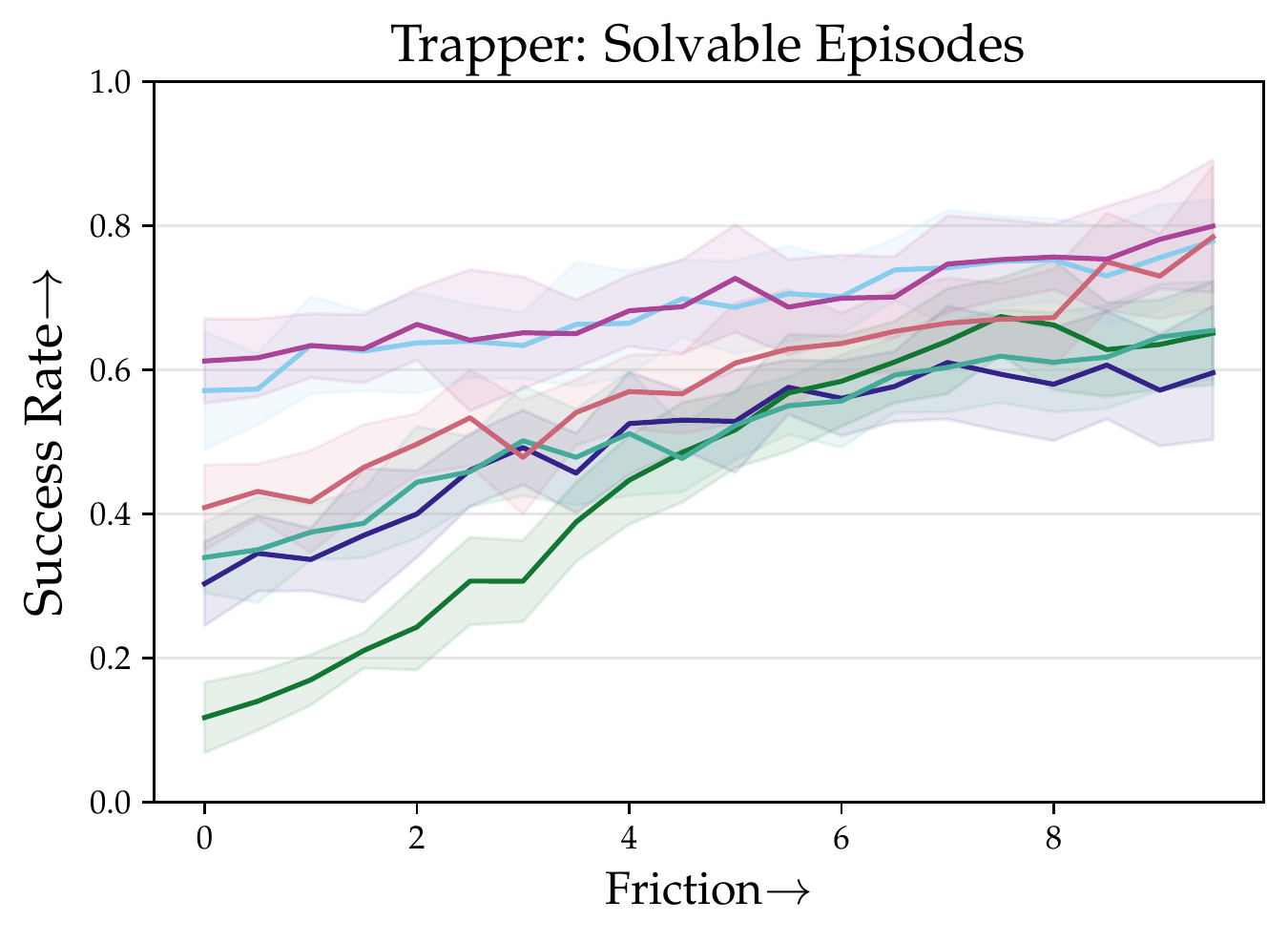}
\includegraphics[ width=.75\linewidth]{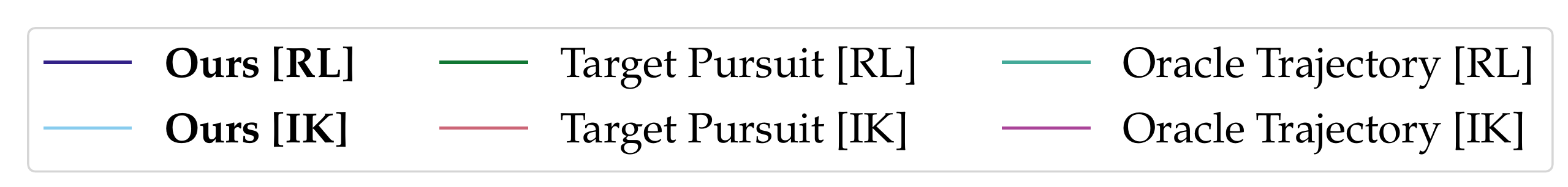} 
\vspace{-0.1in}
    \caption{Success rates for our Trapper-Dynamic environment. On the top we show the results for initial velocities between $0$ and $5$. On the bottom we show results for friction loss between $0$ and $10$. On the right are the success rates for both experiments for episodes we deem solvable. On the left are the success rates for the experiments over all episodes.}
    \label{fig:trapper_figs}
    \vspace{-0.25in}
\end{figure}
\vspace{-0.075in}
\subsection{Results on Trapper-Dynamic}
Our results are visualized in Figure~\ref{fig:trapper_figs}. Except for cases where the ball barely moves (velocity $<1.5$ or friction $> 5$), both our models outperform the target pursuit baseline by about $20\%$. In the target pursuit approach, the arm trails behind the ball, and because the arm moves slower than the ball, it never catches up. Our approach accounts for this by moving towards a future location, saving time that might have been spent following the ball. This allows it to reach certain locations before the ball. As the ball gets faster, the value of $~\tilde{g}$ has to be more accurate; this is why we see our success rate fall when we increase speed or decrease friction.

The results of `oracle trajectory' and our predicted trajectory are similar across both the RL and the IK static planner, except for high-velocity episodes utilizing the IK planner where using the `oracle' trajectory is better. This is potentially because the inverse kinematics solvers are more exact and can benefit from very precise trajectories. However, since the results are still relatively close, we can infer that accurately synchronizing the arm's timing with the ball is the difficult part of the problem. In other words, noise in the trajectory prediction (or $~\tilde{g}$) is not very significant as long as the overall predicted direction is correct.
\vspace{-0.075in}
\subsection{Results on Dynamic FetchReach}
To compare our approach to a recent state-of-the-art reinforcement learning algorithm, we run our preliminary set of experiments on a dynamic version of the FetchReach environment. This is one of the environments used by~\cite{fang2018dher} to evaluate Dynamic Hindsight Experience Replay (DHER). This task is much easier than Trapper-Dynamic because there is no friction, and there are no collisions. As can be seen in Figure~\ref{fig:rest_figs}(right), our approach significantly outperforms~\cite{fang2018dher}. At the largest gap, our approach outperforms it by almost $60\%$. This demonstrates that our approach can easily be successfully applied to other environments. Note that we were unable to get any recent state-of-the-art approach for dynamic tasks work on our Trapper-Dynamic task.
\vspace{-0.075in}
\subsection{Ablation Analysis}
\subsubsection{RL vs.\ IK Static Planners} We evaluate the differences between the RL and the IK static planners for our task. Given our current setup, the IK planner outperforms the RL planner by $\sim$20\%. The key challenge for solving Trapper using RL was finding a static policy that could change its target location mid episode. We anticipate that the RL policy not being agile enough to change target locations explains a large percentage of the gap between the two planners. However, we still see value in experimenting with the RL planner. One advantage is speed, the RL planner takes $<$1ms to find the next action, while the IK planner takes $\sim$15ms. Additionally, the RL planner is better able to recover from a variety of failure modes. E.g., if the box hits the table and/or misses the ball, the RL planner has learned to lift the arm and try again, but the IK planner is unable to recover.

\begin{figure}[t]
\vspace{0.05in}
    \centering
    \includegraphics[width=.46\linewidth]{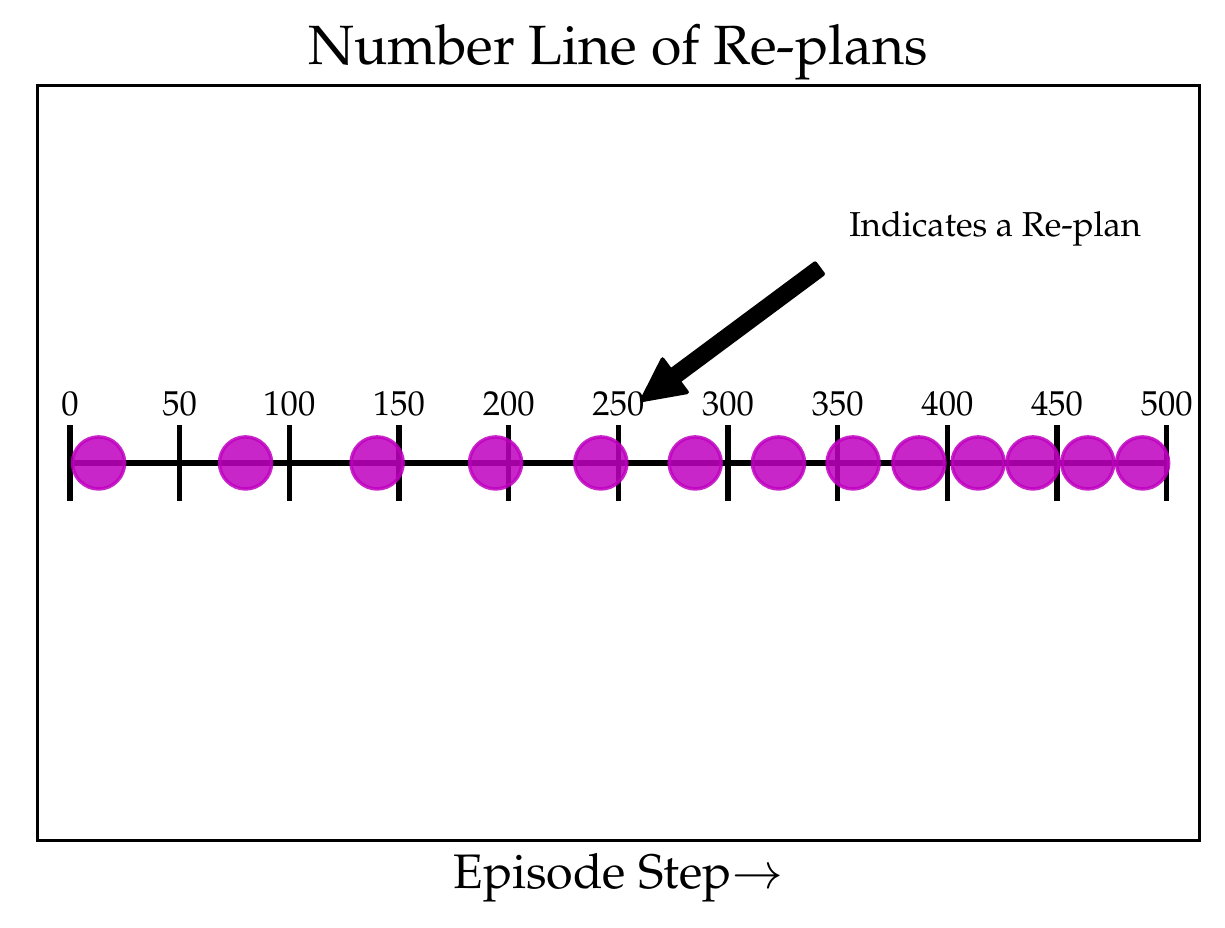} \hfill
\includegraphics[width=0.48\linewidth]{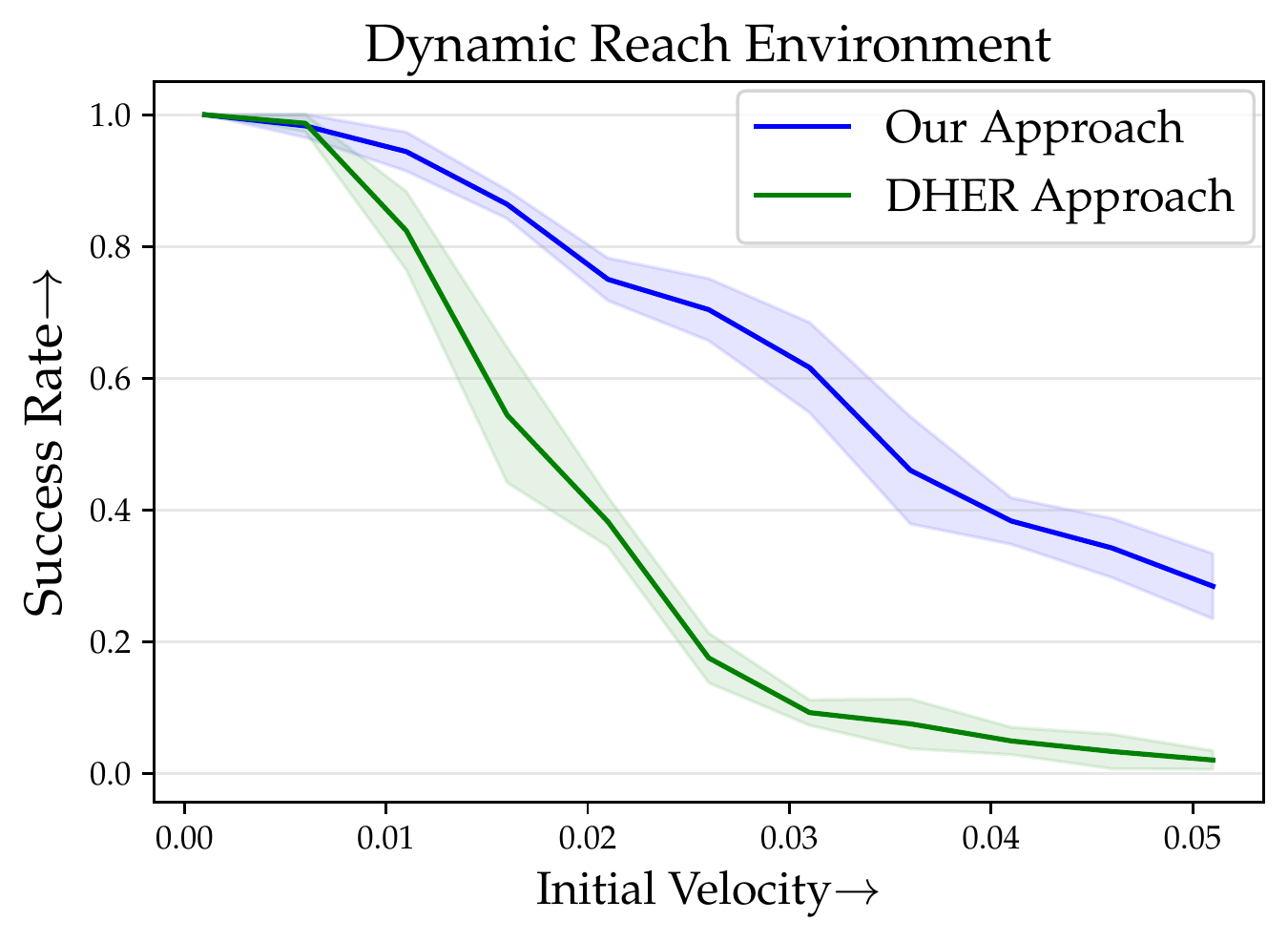}
\vspace{-0.12in}
    \caption{\textbf{(left):} A visualization of when re-plans occur during an episode. \textbf{(right):} Success rate of our algorithm as well as the DHER \cite{fang2018dher} algorithm on a dynamic version of the FetchReach environment from OpenAI Gym \cite{DBLP:journals/corr/AndrychowiczWRS17}. We evaluate both methods for different velocities of the ball.}
    \label{fig:rest_figs}
    \vspace{-0.25in}
\end{figure}

\subsubsection{Timing Analysis}
Since the entire episode occurs in $<2$sec, the timing of the entire planning process is important. When the re-plan strategy is fixed, it leads to 13 re-plans over the course of a 500-step episode (visualized in Figure~\ref{fig:rest_figs}(left)).
On average the intermediate goal search takes 35ms running on 1 GPU, which includes 4 forwards passes of the ETA network, 1 forward pass of the Trajectory Network, and all of the data processing required to find the target location. Given these numbers, we can see that our algorithm uses an appropriate number of re-plans and that re-planning at each time step is too slow for this task.

\section{CONCLUSION}
\label{sec:conclusion}
We presented an approach to adapt static models to work with dynamic targets. Our approach is intuitive, easy to implement, and can easily be adapted to newer environments. We showed that we could synchronize the timing between a moving target object and a moving robot, which has has not been widely studied in learning-based robotics. Results demonstrate that the approach generalizes to different environment parameters and that our approach can work with noisy future predictions.

\noindent\textbf{Acknowledgements:} This project was partially funded by DARPA SAIL-ON program (W911NF2020009). AS would like to thank James Davidson and Rahul Sukthankar for helpful early discussions.

\bibliographystyle{IEEEtran}
\bibliography{main}

\end{document}